# Identification of physical processes via combined data-driven and data-assimilation methods

Haibin Chang and Dongxiao Zhang *

ERE, BIC-ESAT, and SKLTCS, College of Engineering, Peking University, Beijing 100871, China.

*Corresponding author. Email: dxz@pku.edu.cn

**Abstract**
With the advent of modern data collection and storage technologies, data-driven approaches have been developed for discovering the governing partial differential equations (PDE) of physical problems. However, in the extant works the model parameters in the equations are either assumed to be known or have a linear dependency. Therefore, most of the realistic physical processes cannot be identified with the current data-driven PDE discovery approaches. In this study, an innovative framework is developed that combines data-driven and data-assimilation methods for simultaneously identifying physical processes and inferring model parameters. Spatiotemporal measurement data are first divided into a training data set and a testing data set. Using the training data set, a data-driven method is developed to learn the governing equation of the considered physical problem by identifying the occurred (or dominated) processes and selecting the proper empirical model. Through introducing a prediction error of the learned governing equation for the testing data set, a data-assimilation method is devised to estimate the uncertain model parameters of the selected empirical model. For the contaminant transport problem investigated, the results demonstrate that the proposed method can adequately identify the considered physical processes via concurrently discovering the corresponding governing equations and inferring uncertain parameters of nonlinear models, even in the presence of measurement errors. This work helps to broaden the applicable area of the research of data driven discovery of governing equations of physical problems.

## Introduction

Physical problems may consist of several processes. For example, for contaminant solute transport in subsurface formation, simultaneous processes may exist, such as advection, dispersion, and reaction (7); hydraulic fracturing in oil/gas reservoirs is possible because of the coupled processes of fluid flow, geomechanics, and heat transfer (*11*); and the viability of geological sequestration of carbon dioxide depends on the collective effects of hydrodynamic trapping, residual trapping, dissolution trapping, gravitational instability, and mineral trapping, while the dominant processes may vary during the course of storage (*22*). Modeling such processes is important for characterizing corresponding physical problems. First principle derivation by resorting to conservation law can lead to rigorous models for some physical processes. However, it may not be applicable to some complex processes, for which approximate or empirical models are usually proposed based on laboratory experiments or data analyses. Alternative models of various accuracies or complexities may be proposed for the same physical process by considering different conditions, and these models usually have several parameters. For the specific occurrence of a physical problem, which processes occurred (or dominated) may be unclear. Moreover, which is the proper empirical model among the alternatives for a specific process? What are the values of the model parameters of the empirical model? Addressing these questions can be summarized into two tasks: determining the governing equation, which includes identifying the occurred (or dominated) processes and selecting the proper empirical model(s); and estimating the uncertain model parameters in the governing equation. Developing a methodology for efficiently solving these is crucial for many physical problems.

When the governing equation and its solving scheme of an investigated problem is known, estimating the uncertain model parameters can be implemented by utilizing data-assimilation (inverse modeling) methods. In data-assimilation methods, the uncertain model parameters can be automatically updated to maximize the posterior probability density function. Various data-assimilation methods exist, such as the Markov chain Monte Carlo method (*4, 8, 10*), the gradient-based method (*13*), and the ensemble-based method (*3, 5, 6, 9*). The prerequisite of implementing data-assimilation methods is that the model of the investigated problem that includes the governing equation and its solving scheme is known. If some of the occurred (or dominated) physical processes are not clear, the proper empirical model for a specific process cannot be determined, or the constitutive relations of some quantities are not explicitly known, the model will be vague, which will impede the implementation of data-assimilation method.

For these physical problems, determining the governing equation, which includes identifying the occurred (or dominated) processes and selecting the proper empirical model, is critical. Usually, this can be accomplished by combining theoretical derivations and laboratory experiments. However, this may be time-consuming and present some technical obstacles. When spatiotemporal measurements are available, the data-driven method may provide a viable option for discovering the governing equation of a physical problem. Several recent works exist in investigating data-driven discovery of dynamical systems (*1*, *12*, *14*, *19*, *20*) and partial differential equations (PDE) (*17*, *18*). In these works, sparse regression, such as sequential threshold ridge regression and the least absolute shrinkage and selection operator (LASSO), constitutes an essential technique for identifying the terms in the governing equation from a large candidate library. The Gaussian process technique is also utilized for data-driven discovery of differential equations in some works (*15*, *16*).

In the extant literature about learning PDE, the general form of the investigated PDE takes the form $\partial u / \partial t = \Phi(u)\alpha$, where $u$ is the response of the physical problem, $\Phi(u)$ is the candidate term library, and $\alpha$ is the coefficient. For the physical problem considered in this work, because the empirical model usually contains model parameters in a nonlinear manner, the PDE takes the form $\partial u / \partial t = \Phi(u, m)\alpha$, where $m$ is the uncertain model parameter that cannot be included in $\alpha$. Here, note that the empirical model is not only used for modeling a physical process, but also for modeling the constitutive relationship between variables of many physical problems. For example, for modeling flow in unsaturated porous media, empirical models are needed to model the constitutive relationships between conductivity, water content, and hydraulic head (*21*). Thus, this generalized form of PDE can cover a wider range of physical problems. However, such an extension cannot be resolved with the current data-driven PDE discovery approaches. In this study, in order to identify the considered physical processes from data, an innovative framework that combines the data-driven and data-assimilation method is proposed. For illustrating the proposed method, we focus here on the particular problem of contaminant solute transport in subsurface formation, while the proposed method can be easily applied to numerous other problems in different fields. This work helps to broaden the applicable area of the research of data-driven discovery of governing equations of physical problems.

## Results

### Identification of solute transport processes

Here, we evaluate the proposed method by considering the problem of contaminant solute transport in subsurface formation. Solute transport in subsurface formation may be subject to a number of different processes (*7*). For a site with solute transport, our goal is to use spatiotemporal solute concentration measurements to identify the occurred (or dominated) processes, select the proper empirical model for describing a reaction, and estimate the model parameters. This can be generalized as identifying PDE with the form:

$$\frac{\partial u}{\partial t} = \Phi(u,m)\alpha, \tag{1}$$

where $u$ denotes the response of a physical problem; $\Phi(u,m)$ denotes the library of candidate processes and empirical models; $\alpha$ denotes the coefficient; and $m$ denotes the model parameters of the empirical model that cannot be included in $\alpha$.

For building the candidate library of the solute transport problem, three processes, advection (ADV), dispersion (DIS), and sorption (SORP), are considered in this work. Here, it is assumed that prior knowledge exists about the modeling of each considered candidate process. For the change in solute concentration with time, $\partial C/\partial t$, where $C$ denotes the concentration of solute in aqueous phase, one-dimensional ADV and DIS are modeled by $-v_x \partial C/\partial x$ and $D_L \partial^2 C/\partial x^2$, respectively, where $v_x$ denotes the average linear groundwater velocity, and $D_L$ denotes the longitudinal dispersion coefficient. The model of SORP can be expressed by $-(\rho_b/\theta)\partial C^*/\partial t$, where $\rho_b$ denotes the bulk density of the aquifer, $\theta$ denotes the porosity, and $C^*$ denotes the amount of solute adsorbed per unit weight of solid. Different from the ADV and DIS, the SORP may not be easily modeled by first principle derivation. Different empirical models are usually proposed (based on laboratory experiments) for modeling SORP by considering different conditions. Two equilibrium sorption models, Freundlich sorption isotherm (F-SORP) and Langmuir sorption isotherm (L-SORP), are considered here, which are expressed as $C^* = K_f C^a$ and $C^* = (K_l \bar{S} C)/(1+K_l C)$, respectively, where $K_f$ denotes the Freundlich constant, $a$ denotes the Freundlich exponent, $K_l$ denotes the Langmuir constant, and $\bar{S}$ denotes the total concentration of sorption sites available. Empirical models normally contain model parameters, which are traditionally obtained by data fitting. For the considered solute transport, the candidate library can be expressed as:

$$\Phi(C,m) = \left[ \frac{\partial C}{\partial x}, \frac{\partial^2 C}{\partial x^2}, C^{a-1}\frac{\partial C}{\partial t}, \frac{1}{(1+K_l C)^2}\frac{\partial C}{\partial t} \right] \qquad (2)$$

and,

$$m = (a, K_l) \qquad (3)$$

Here, note that the four terms of $\Phi(C,m)$ shown in Eq. 2 are used for denoting ADV, DIS, F-SORP, and L-SORP, respectively, and some parameters of the process models ($-v_x$, $D_L$, $-(\rho_b/\theta)aK_f$, and $-(\rho_b/\theta)K_l \overline{S}$) are included in $\alpha$ while the nonlinear terms $C^{a-1}$ and $1/(1+K_l C)^2$ cannot be included in $\alpha$.

For proof of the concept, the data used in this work are obtained from numerical simulation. A benchmark problem of solute transport from MT3DMS software (*23*) is adopted here. We briefly describe the problem setup here, and additional details can be found in the Supplementary Materials. It is a one-dimensional problem. The flow field is steady state. A solute is released at the origin for 160 s. The solute concentration is measured from 300 s to 1100 s at 101 evenly-distributed spatial locations. The measurement interval is 0.5 s. Three solute transport scenarios are designed here: scenario 1 contains the two processes of ADV and DIS; scenario 2 contains the three processes of ADV, DIS, and F-SORP; and scenario 3 contains the three processes of ADV, DIS, and L-SORP. For the three scenarios, $v_x$ and dispersivity are set as 0.01 and 1, respectively. For scenario 2, $K_f$ and $a$ are set as 0.05 and 0.7, respectively. For scenario 3, $K_l$ and $\overline{S}$ are set as 100 and 0.003, respectively. The three scenarios are particularly designed for testing the performance of the proposed method for identifying the occurred (or dominated) processes, selecting the proper empirical model, and estimating the model parameters. MT3DMS software is utilized for running the simulations to obtain the measurement data for the three scenarios.

Because *m* is usually uncertain, the traditional data-driven method cannot be utilized for learning the equation shown in Eq. 1. In order to learn this type of equation, a combined data-driven and data-assimilation method is proposed. For implementing the proposed method, the data are first divided into a training data set and a testing data set. In this work, the data are divided according to time sequence. The first 60% of data are used as a training data set, and the remaining 40% of data are used as a testing data set. The prior distribution of *m* is needed to be prescribed according to prior knowledge.

Here, *a* is supposed to follow a uniform distribution and take values from 0.25 to 0.75, that is, $a \in U[0.25, 0.75]$; and $K_l$ is supposed to follow a uniform distribution and take values from 30 to 150, that is, $K_l \in U[30, 150]$. Implementing the proposed method is started with an initial sample of *m*. Next, the corresponding equation is learned using the training data set, and the prediction error of the learned equation for the testing data set is calculated. Then, *m* is updated using the data-assimilation method. Parameter updating and equation learning are then continued until a convergence criterion is met. Additional details of the proposed method can be found in the Methods section.

Starting with 100 initial samples of *m*, we obtain 100 groups of results independently. Fig. 1 shows the results of identification of three solute transport scenarios, which include the learned coefficients of the normalized candidate terms (Figs. 1a-1c), the estimation of *a* (Figs. 1d-1f), and the estimation of $K_l$ (Figs. 1g-1i). Since 100 groups of results are obtained, the coefficients of normalized candidate terms are shown in a box plot. From Fig. 1a, one can see that, for scenario 1, ADV and DIS are identified to be the occurred (or dominated) processes, while the coefficients of the normalized two empirical models of SORP are close to zero, indicating that for this scenario, SORP can be ignored. From Fig. 1b, one can see that, for scenario 2, the three candidate processes are all identified to be occurred, and it is clear that F-SORP is the proper empirical model for SORP by observing that the coefficient of normalized L-SORP is close to zero. Similarly, from Fig. 1c, one can see that, for scenario 3, the three candidate processes are all identified to be occurred, and it is clear that the L-SORP is the proper empirical model for SORP by observing that the coefficient of normalized F-SORP is close to zero. These results demonstrate that the proposed method can identify the occurred (or dominated) processes and select the proper empirical model well. In the proposed method, a gradient-based data-assimilation method is utilized for updating *m*. If a candidate empirical model is identified to be not occurred (or dominated), the associated model parameters will have a near zero (or small) gradient, and thus there will be no (or slight) update of these parameters. In contrast, the associated model parameters will be updated if the corresponding candidate empirical model is identified to be occurred (or dominated). This is verified by the results shown in Figs. 1d-1i. Since F-SORP is only occurred in scenario 2, there is no obvious update of its model parameter, *a*, for scenario 1 (as shown in Fig. 1d) and scenario 3 (as shown in Fig. 1f). While, for scenario 2, the updated *a* is close to the true value with a mean of 0.700 and a standard deviation of 0.0031. Similarly, there is no obvious update of $K_l$ for scenario 1, and there is slight update of $K_l$ for scenario 2. While, for scenario 3, the updated $K_l$ is close to the true value with a mean of 100.349 and a standard deviation of 0.086. These results show that the proposed method can estimate the model parameters of the selected empirical model well. For identification of physical processes, the proposed method can be implemented only one time with one initial sample of $m$. Implementing the proposed method multiple times with different initial samples of $m$ can assist to avoid getting stuck at the local minimums of the objective function when updating $m$ using the data-assimilation method. By observing the results shown in Fig. 1, the final learned coefficients of the normalized candidate terms and updated model parameters of the selected empirical model from all of the implementations have small variability, which may indicate that the objective function of the considered cases only have a global minimum and all of the results converged to it. Thus, for the case considered here, it may only be necessary to implement the proposed method one time with one initial sample of $m$. However, if there is no prior information about the shape of the objective function with respect to model parameters, multiple implementations are recommended. For determining the learned equation, the normalized terms that have coefficients close to zero will be discarded in order to obtain a parsimonious model. The final learned equations of the three scenarios are given in Table 1. For the coefficients of the PDE terms, we use the mean of the results of all of the implementations. By comparison with the true PDE, the learned PDE has high accuracy for both the coefficients and model parameters.

Other tests with different parameter setups are implemented, and the results are also satisfactory (see Supplementary Materials).

**The influence of data noise**

In previous cases, the data utilized are clean, i.e., there is no measurement error. However, in reality, the measurement data may be associated with some noise. Here, we test the performance of the proposed method with noisy data. We synthetically add noise to data as:

$$C(x,t) = C(x,t) \times (1+\delta \times e) \tag{4}$$

where $\delta$ denotes the noise level; and $e$ denotes the uniform random variable taking values from -1 to 1. Three noise levels, 1%, 5%, and 10%, are added to the data of the three solute transport scenarios. For $\Phi(C,m)$ shown in Eq. 2, the spatial derivatives up to order two and temporal derivative need be calculated using spatiotemporal data.

Since finite difference is utilized for calculating derivatives in this work, small data noise may incur huge error in the calculated derivatives. Thus, pretreatment of the noisy data is requisite. In this work, the polynomial technique is utilized to smooth the noisy data. Additional details about the pretreatment can be found in the Supplementary Materials.

Fig. 2 presents the results of the learned coefficients of the normalized terms with three data-noise levels. From the figures, it can be seen that, for scenario 1 and scenario 3, the occurred (or dominated) processes and the proper empirical model can be satisfactorily identified by pretreating the noisy data. For 10% noise, however, there are some implementations with spurious results. Moreover, for scenario 2, the learned coefficients exhibit larger variability than that of scenario 1 and scenario 3. The coefficient of the normalized L-SORP is not as small as that with clean data. By comparing the magnitude of the absolute value, the coefficient of normalized F-SORP is larger than that of L-SORP for the three noise levels. However, as the noise level becomes larger, the superiority of F-SORP becomes smaller. Since only one model is needed for SORP, F-SORP can be identified as the better empirical model for SORP under the three noise levels. The results of the updated model parameter with three data-noise levels are given in the Supplementary Materials. Overall, as noise level becomes larger, accuracy decreases. It is also seen that data noise may incur larger variability in the results from different implementations, which will increase the difficulty for physical process identification and model parameter inference. Considering that a prediction error of the learned model for the testing data will be calculated in the proposed method, the abnormal results of some implementations may be screened out via analyzing the prediction errors of the final learned equations. By screening out the abnormal results, the variability of the learned $\alpha$ and updated *m* will decrease. On the other hand, in order to improve the accuracy of the results of $\alpha$ and *m*, the proposed method may be implemented again with a modified candidate library, which only retains the identified terms from the first implementation. The learned equations with three data-noise levels are given in Table 1. Similarly, as noise level becomes larger, the accuracy of the learned equation decreases. Additional discussion and results can be found in the Supplementary Materials.

Data noise constitutes a challenge for the proposed method, and further works are requisite.

**Learning equations using the combined method**

In previous cases, concerning prior knowledge about the considered solute transport problem, it is assumed to know that there exist ADV, DIS, and SORP. In reality, the potential physical processes or how they are modeled for a physical problem may be unclear. In order to demonstrate the applicability of the proposed method to more challenging problems, here the prior information about the considered solute transport problem is relaxed as that SORP is regarded as a potential process for solute transport with two possible empirical models, while no other information exists about the other potential processes. In order to learn the governing equation, a new candidate library is designed as:

$$\Phi(C,m) = \left[ C, C^2, \frac{\partial C}{\partial x}, \frac{\partial^2 C}{\partial x^2}, \frac{\partial^3 C}{\partial x^3}, \frac{\partial C^2}{\partial x}, \frac{\partial^2 C^2}{\partial x^2}, \frac{\partial^3 C^2}{\partial x^3}, C^{a-1}\frac{\partial C}{\partial t}, \frac{1}{(1+K_l C)^2}\frac{\partial C}{\partial t} \right]. \quad (5)$$

Besides the last two terms, which are used for modelling F-SORP and L-SORP, respectively, there are eight other terms utilized for determining other physical processes. The data of the three previously-designed scenarios are also utilized here for learning equation. The prior distributions of $a$ and $K_l$ are the same as previous cases.

Fig. 3 shows the results of the learned coefficients of the normalized terms and the estimated model parameters for the three scenarios. From Figs. 3a-3c, one can see that the correct terms can be selected for the three scenarios. Regarding the estimation of model parameters, one can see that there is no update of $a$ and $K_l$ for scenario 1. The updated $a$ for scenario 2 has a mean of 0.701 and a standard deviation of 0.022. The updated $K_l$ for scenario 3 has a mean of 101.025 and a standard deviation of 0.388.

While, for some implementations, there is large update of $K_l$ for scenario 2 and $a$ for scenario 3. This may be caused by the error of the calculated gradient. Since the empirical model can be correctly selected, this spurious update of model parameter will not influence the final learned equation. For determining the learned equation, the normalized terms that have coefficients close to zero will be discarded. In order to further improve the accuracy of the learned equation, a new candidate library can be built by deleting the unnecessary terms. Choosing the selected terms of the three scenarios, the new candidate library will be the same as that shown in Eq. 2. Implementing the proposed method again with the new candidate library, the same learned equations as those of previous cases would be obtained.

From the results here, it can be seen that the proposed method can be utilized for learning the equation with the form shown in Eq. 1.

## Discussion

In recent works that investigate data-driven discovery of dynamical systems and PDEs, the aim is to learn one set of parameters, i.e., the $\alpha$ utilized in this work. For physical problem, it may contain physical process that needs to be modeled by empirical models or augmented with constitutive relations. In addition, the empirical model usually contains model parameters, *m*, which cannot be included in $\alpha$. Thus, data-driven discovery of this kind of physical problem comprises two tasks, which are to learn $\alpha$ and to estimate *m*. Through focusing on a particular problem of contaminant solute transport in subsurface formation, the results demonstrate that the proposed method that combines the data-driven and data-assimilation methods can solve this problem well. The proposed method provides an efficient option for identifying occurred (or dominated) physical processes, selecting proper empirical model, and estimating model parameters. Generally, the proposed method can be utilized to learn the equation with the form shown in Eq. 1. Considering that empirical model is not only used for modeling a physical process, but also for modeling constitutive relationships between variables of many physical problems, this form of PDE can cover a wider range of physical problems. This study assists to broaden the applicable area of the research of data-driven discovery of governing equations of physical problems.

In order to learn the appropriate terms/models ($\alpha$) and estimate the model parameters (*m*), the proposed method can be implemented only one time with one initial sample of *m*. Since the gradient-based data-assimilation method is utilized, multiple implementations with different initial samples of *m* are recommended for avoiding getting stuck at the local minimums of the objective function. For determining the final learned $\alpha$ and estimated *m*, it is helpful to screen out spurious results of some implementations based on analyzing the prediction errors. In this work, for case studies, 100 implementations are performed to test the performance of the proposed method, while it is not necessary to perform this large number of implementations. The results demonstrate that a small number of implementations may be sufficient for determining the learned equation.

Traditionally, implementing the data-assimilation method requires forward simulations for predicting model responses with respect to given model parameters. In this work, however, the prediction error introduced in the proposed data-assimilation method is calculated using a testing data set, and no numerical simulations of the forward problem are revolved. Thus, the data-assimilation method utilized in this work can also be understood as a data-driven method.

Since spatial and temporal derivatives need to be calculated in the proposed method, data noise presents a challenge. The results show that, for larger noise, the accuracy of the learned $\alpha$ and estimated $m$ will decrease. After obtaining the learned PDE, in order to improve the accuracy of the coefficients and parameters of the PDE terms, a numerical scheme of the learned PDE may be developed. Moreover, a data-assimilation method that uses the developed numerical scheme for performing forward simulations may be implemented to further update the coefficients and parameters of the PDE terms.

Combing the data-driven and data-assimilation methods constitutes a viable strategy for solving the discussed problem. Besides the methods utilized in this work, other data-driven and data-assimilation methods may be used as alternatives. Especially, for data-driven methods, sparsity constrained methods, such as LASSO, may be employed to replace the least square regression that is utilized in this work to improve the performance of the proposed method.

Some physical problems are described by coupled equations, where empirical models of constitutive relationships can be included as supplementary equations. The proposed method is amenable to these kinds of problems.

## Methods

Identifying the kinds of physical processes discussed in this work can be understood as learning the PDE with the form shown in Eq. 1. In order to achieve this, spatiotemporal measurement data are divided into a training data set with number $n_{train}$ and a testing data set with number $n_{test}$. For the training data set, Eq. 1 can be rewritten as:

$$\frac{\partial U}{\partial t} = \Phi(U,m)\alpha, \tag{6}$$

where $U = [u_1, u_2, ..., u_{n_{train}}]^T$; and $u_i$ denotes the $i^{th}$ training data. For any $m$, the corresponding coefficient $\alpha$ is learned by least square regression as:

$$\alpha = \alpha(m) = \left(\Phi(U,m)^T \Phi(U,m)\right)^T \Phi(U,m)^T \frac{\partial U}{\partial t}. \tag{7}$$

For the learned equation, a prediction error $\varepsilon$ is calculated using the testing data set as:

$$\varepsilon = \varepsilon(m) = \sum_{n=1}^{n_{test}} \left(\frac{\partial u_n}{\partial t} - \Phi(u_n,m)\alpha(m)\right)^2. \tag{8}$$

In order to update $m$, an objective function is defined as:

$$O(m) = \frac{1}{2}(\varepsilon(m) - \varepsilon^{obs})^T C_\varepsilon^{-1}(\varepsilon(m) - \varepsilon^{obs}) + \frac{1}{2}(m - m^{pr})^T C_M^{-1}(m - m^{pr}). \tag{9}$$

where $\varepsilon^{obs}$ denotes the observed (or target) prediction error; $m^{pr}$ denotes the initial sample of *m*; $C_\varepsilon$ denotes the covariance of the prediction error; and $C_M$ denotes the covariance of *m*. Supposing that no error will be incurred in calculating the partial derivatives in Eq. 8, the true PDE will have no prediction error, that is, $\varepsilon^{obs} = 0$.

The data-assimilation method utilized in this work for updating *m* takes the form (*2*):

$$m_{l+1} = m_l \\
- \frac{1}{1+\lambda_l}\left[C_M - C_M G_l^T \left((1+\lambda_l)C_\varepsilon + G_l C_M G_l^T\right)^{-1} G_l C_M\right] C_M^{-1}\left(m_l - m^{pr}\right) \quad (10) \\
- C_M G_l^T \left((1+\lambda_l)C_\varepsilon + G_l C_M G_l^T\right)^{-1} \left(\varepsilon(m_l) - \varepsilon^{obs}\right),$$

where *l* denotes the iteration index; $\lambda$ denotes the multiplier; and *G* denotes the gradient. In this work, *G* is calculated by finite difference.

In summary, starting with one initial sample of *m*, $\alpha$ will be learned and $\varepsilon$ will be calculated. Then, the data-assimilation method can iteratively update *m* to minimize an objective function. After each update of *m*, $\alpha$ will be relearned and $\varepsilon$ recalculated. This process will continue until a convergence criterion is met. Additional details of the proposed method can be found in the Supplementary Materials.

## Acknowledgments

This work is partially funded by the National Natural Science Foundation of China (Grant No. U1663208 and 51520105005), and the National Science and Technology Major Project of China (Grant No. 2017ZX05009-005 and 2016ZX05037-003).

# Figures and Tables

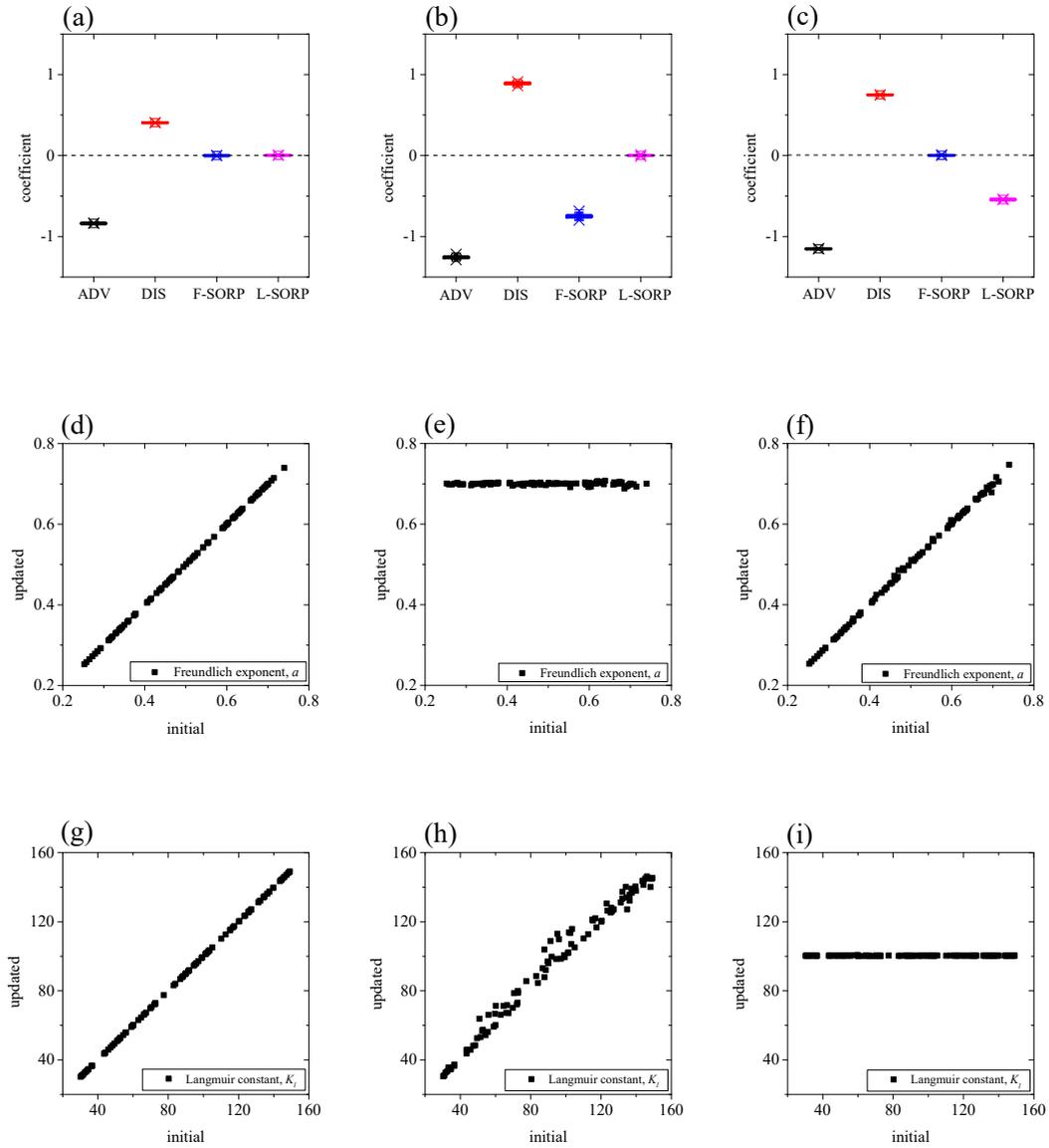

**Fig. 1. The results of identification of three solute transport scenarios.** The learned coefficients of normalized terms of scenario 1 (**a**), scenario 2 (**b**), and scenario 3 (**c**). The estimation of Freundlich exponent of scenario 1 (**d**), scenario 2 (**e**), and scenario 3 (**f**). The estimation of Langmuir constant of scenario 1 (**g**), scenario 2 (**h**), and scenario 3 (**i**).

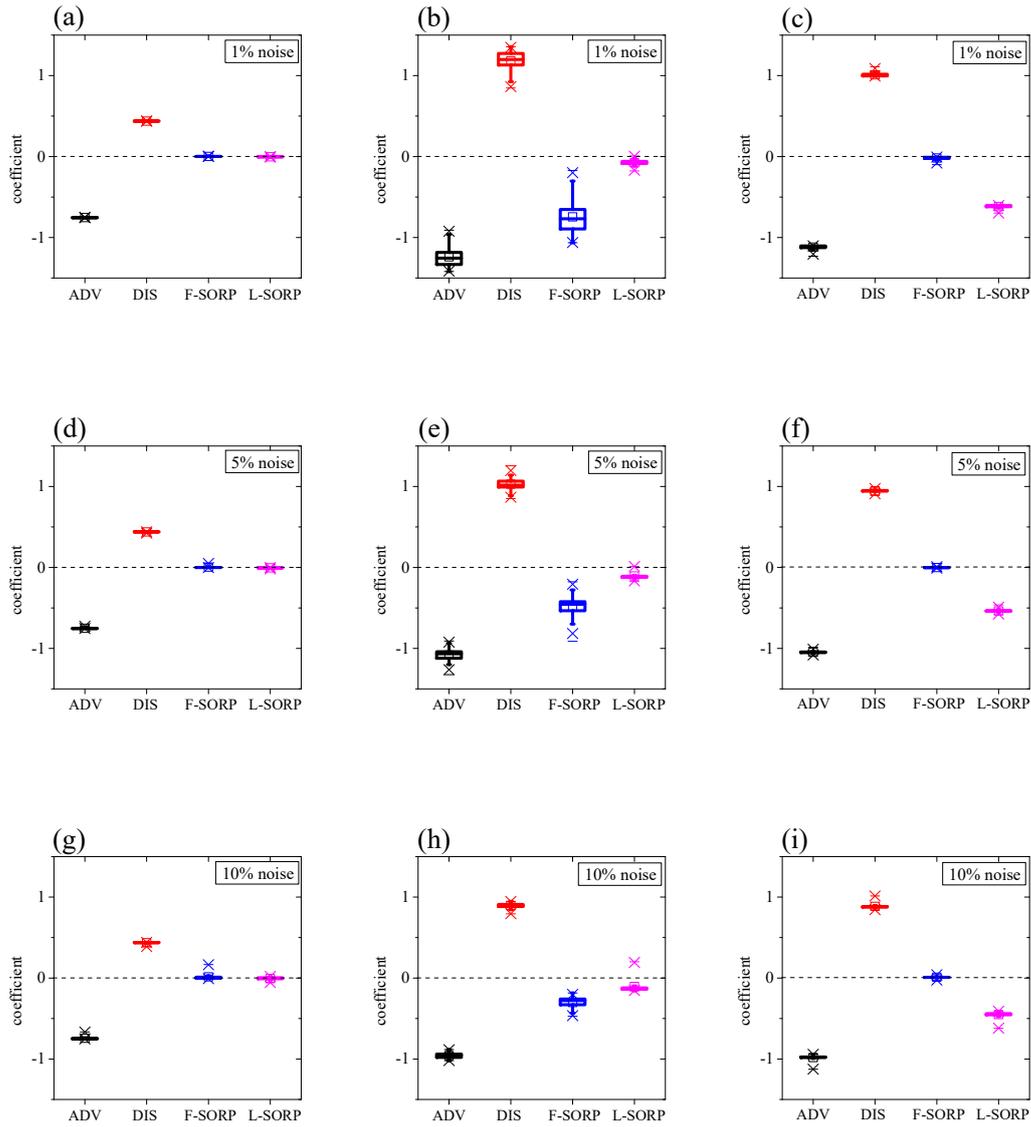

**Fig. 2. The learned coefficients of normalized terms with three data-noise levels.**
With 1% noise, the results of scenario 1 (**a**), scenario 2 (**b**), and scenario 3 (**c**). With 5% noise, the results of scenario 1 (**d**), scenario 2 (**e**), and scenario 3 (**f**). With 10% noise, the results of scenario 1 (**g**), scenario 2 (**h**), and scenario 3 (**i**).

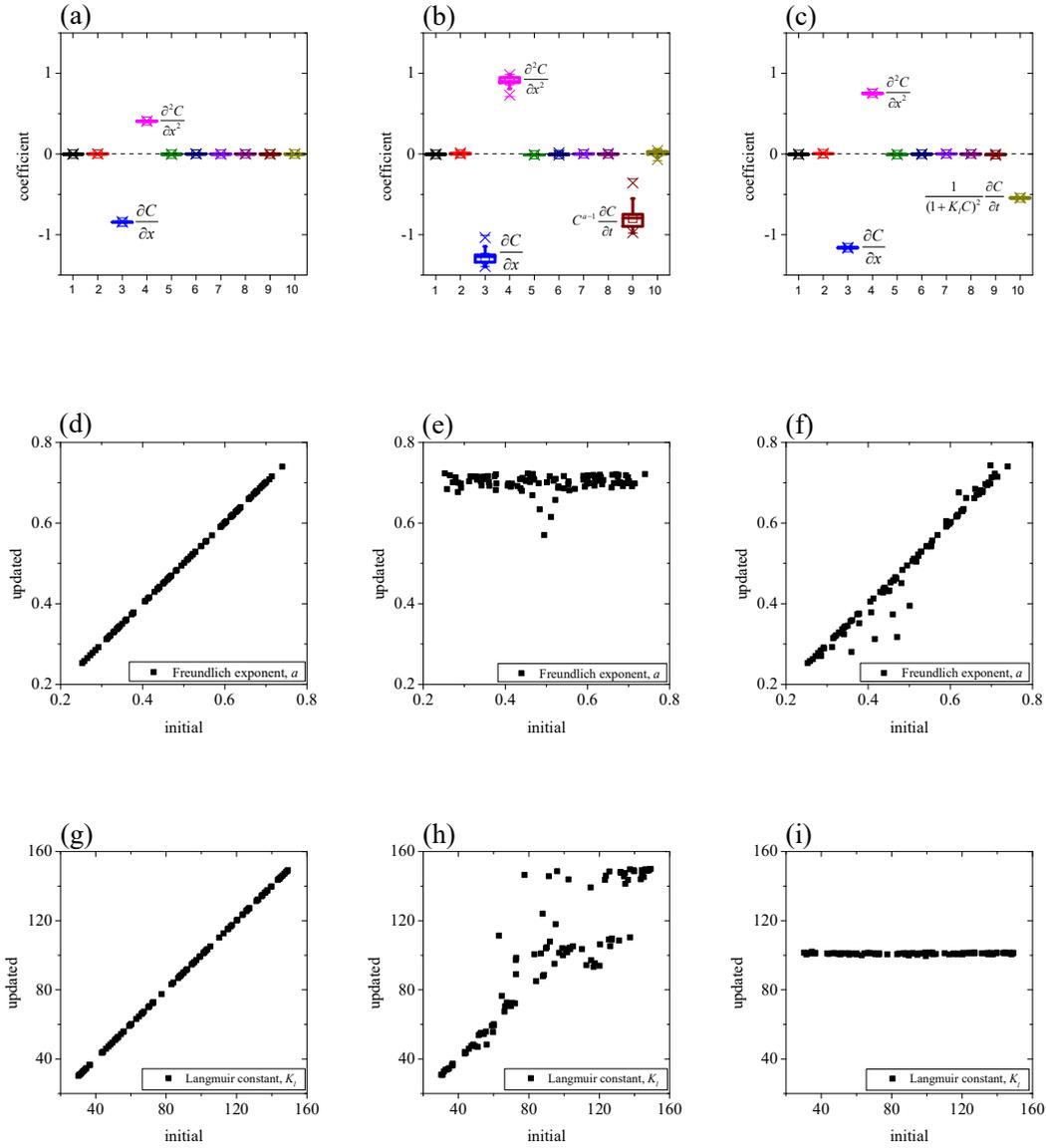

**Fig. 3. The results of learning equations of three solute transport scenarios.** The learned coefficients of normalized terms of scenario 1 (**a**), scenario 2 (**b**), and scenario 3 (**c**). The estimation of Freundlich exponent of scenario 1 (**d**), scenario 2 (**e**), and scenario 3 (**f**). The estimation of Langmuir constant of scenario 1 (**g**), scenario 2 (**h**), and scenario 3 (**i**).

**Table 1.** The learned equations of solute transport.

| Physical processes | Learned equation |
|---|---|
| ADV and DIS<br><br>$\frac{\partial C}{\partial t} = -0.01\frac{\partial C}{\partial x} + 0.01\frac{\partial^2 C}{\partial x^2}$ | $\frac{\partial C}{\partial t} = -0.00993\frac{\partial C}{\partial x} + 0.00997\frac{\partial^2 C}{\partial x^2}$ (clean data)<br><br>$\frac{\partial C}{\partial t} = -0.00997\frac{\partial C}{\partial x} + 0.01002\frac{\partial^2 C}{\partial x^2}$ (1% noise)<br><br>$\frac{\partial C}{\partial t} = -0.00997\frac{\partial C}{\partial x} + 0.01006\frac{\partial^2 C}{\partial x^2}$ (5% noise)<br><br>$\frac{\partial C}{\partial t} = -0.00991\frac{\partial C}{\partial x} + 0.01004\frac{\partial^2 C}{\partial x^2}$ (10% noise) |
| ADV, DIS, and F-SORP<br><br>$\frac{\partial C}{\partial t} = -0.01\frac{\partial C}{\partial x} + 0.01\frac{\partial^2 C}{\partial x^2}$<br>$- 0.150 C^{0.7-1}\frac{\partial C}{\partial t}$ | $\frac{\partial C}{\partial t} = -0.00989\frac{\partial C}{\partial x} + 0.00992\frac{\partial^2 C}{\partial x^2} - 0.149 C^{0.700-1}\frac{\partial C}{\partial t}$ (clean data)<br><br>$\frac{\partial C}{\partial t} = -0.00967\frac{\partial C}{\partial x} + 0.00968\frac{\partial^2 C}{\partial x^2} - 0.138 C^{0.697-1}\frac{\partial C}{\partial t}$ (1% noise)<br><br>$\frac{\partial C}{\partial t} = -0.00851\frac{\partial C}{\partial x} + 0.00845\frac{\partial^2 C}{\partial x^2} - 0.079 C^{0.657-1}\frac{\partial C}{\partial t}$ (5% noise)<br><br>$\frac{\partial C}{\partial t} = -0.00750\frac{\partial C}{\partial x} + 0.00726\frac{\partial^2 C}{\partial x^2} - 0.037 C^{0.602-1}\frac{\partial C}{\partial t}$ (10% noise) |
| ADV, DIS, and L-SORP<br><br>$\frac{\partial C}{\partial t} = -0.01\frac{\partial C}{\partial x} + 0.01\frac{\partial^2 C}{\partial x^2}$<br>$- 1.287\frac{1}{(1+100.0C)^2}\frac{\partial C}{\partial t}$ | $\frac{\partial C}{\partial t} = -0.00989\frac{\partial C}{\partial x} + 0.00992\frac{\partial^2 C}{\partial x^2} - 1.277\frac{1}{(1+100.349C)^2}\frac{\partial C}{\partial t}$ (clean data)<br><br>$\frac{\partial C}{\partial t} = -0.00976\frac{\partial C}{\partial x} + 0.00982\frac{\partial^2 C}{\partial x^2} - 1.203\frac{1}{(1+100.936C)^2}\frac{\partial C}{\partial t}$ (1% noise)<br><br>$\frac{\partial C}{\partial t} = -0.00921\frac{\partial C}{\partial x} + 0.00929\frac{\partial^2 C}{\partial x^2} - 1.124\frac{1}{(1+114.200C)^2}\frac{\partial C}{\partial t}$ (5% noise)<br><br>$\frac{\partial C}{\partial t} = -0.00857\frac{\partial C}{\partial x} + 0.00863\frac{\partial^2 C}{\partial x^2} - 1.015\frac{1}{(1+136.952C)^2}\frac{\partial C}{\partial t}$ (10% noise) |

# Supplementary Information for

**Identification of physical processes via combined data-driven and data-assimilation methods**

Haibin Chang and Dongxiao Zhang[*]

[*]Corresponding author. Email: dxz@pku.edu.cn

**This PDF file includes:**

**Section S1.** Methods

**Section S2.** Cases

**Fig. S1.** The results of identification of scenario 2 and scenario 3 with different values for $K_f$ and $K_l$, respectively.

**Fig. S2.** The results of identification of scenario 2 and scenario 3 with different value for $v_x$.

**Fig. S3.** The estimation of Freundlich exponent for the three scenarios with three data-noise levels.

**Fig. S4.** The estimation of Langmuir constant for the three scenarios with three data-noise levels.

**Fig. S5.** The prediction error and the learned coefficients of PDE terms for the three scenarios with clean data.

**Fig. S6.** The prediction error and the learned coefficients of normalized terms for scenario 2 with three data-noise levels.

**Fig. S7.** The results of identification of scenario 2 with a modified candidate library.

**Table S1.** The learned equations of scenario 2 and scenario 3 with different values for $K_f$ and $K_l$, respectively.

**Table S2.** The learned equations of scenario 2 and scenario 3 with different value for $v_x$.

**Table S3.** The mean and standard deviation of the coefficients of the PDE terms.

## Section S1. Methods

Identification of the physical processes considered can be understood as identifying the partial differential equation (PDE) with the form:

$$\frac{\partial u}{\partial t} = \Phi(u,m)\alpha, \tag{S1}$$

where $u$ denotes the response of a physical problem; $\Phi(u,m)$ denotes the library of the candidate processes and empirical models; $\alpha$ denotes the coefficient; and $m$ denotes the model parameters of the empirical model that cannot be included in $\alpha$.

In order to learn the PDE shown in Eq. S1, spatiotemporal measurement data are divided into a training data set with number $n_{train}$ and a testing data set with number $n_{test}$. For the training data set, Eq. S1 can be rewritten as:

$$\frac{\partial U}{\partial t} = \Phi(U,m)\alpha, \tag{S2}$$

where $U = [u_1, u_2, ..., u_{n_{train}}]^T$; and $u_i$ denotes the $i^{th}$ training data. Here, $\partial U/\partial t$ is a vector with a size of $n_{train} \times 1$, $\Phi(U,m)$ is a matrix with a size of $n_{train} \times n_{cand}$, and $\alpha$ is a vector with a size of $n_{cand} \times 1$. Here, $n_{cand}$ is used for denoting the number of candidate terms. Because $\Phi(U,m)$ usually contains partial derivatives, calculating derivatives using spatiotemporal data is needed. In this work, finite difference is utilized for calculating derivatives.

For any $m$, the corresponding coefficient $\alpha$ can be learned by least square regression as:

$$\alpha = \alpha(m) = \left(\Phi(U,m)^T \Phi(U,m)\right)^T \Phi(U,m)^T \frac{\partial U}{\partial t}. \tag{S3}$$

Here, note that for learning $\alpha$, $\partial U/\partial t$ and each column of $\Phi(U,m)$ are normalized to have zero mean and unit standard deviation.

Speculating that if the guess of $m$ is far away from the truth, the accuracy of the learned equation will be low. In order to evaluate the learned equation, a prediction error $\varepsilon$ is introduced, which is calculated using the testing data set as:

$$\varepsilon = \varepsilon(m) = \sum_{n=1}^{n_{test}} \left(\frac{\partial u_n}{\partial t} - \Phi(u_n,m)\alpha(m)\right)^2. \tag{S4}$$

Here, note that for calculating $\varepsilon$, we also need to calculate the partial derivatives contained in $\partial u_n/\partial t$ and $\Phi(u_n,m)$ using testing data.

In order to obtain an optimal $m$ using the introduced $\varepsilon$, an objective function is defined as:

$$O(m) = \frac{1}{2}(\varepsilon(m) - \varepsilon^{obs})^T C_\varepsilon^{-1}(\varepsilon(m) - \varepsilon^{obs}) + \frac{1}{2}(m - m^{pr})^T C_M^{-1}(m - m^{pr}), \qquad (S5)$$

where $\varepsilon^{obs}$ denotes the observed (or target) prediction error; $m^{pr}$ denotes the initial sample of $m$; $C_\varepsilon$ denotes the covariance of the prediction error; and $C_M$ denotes the covariance of $m$. The first term on the right-hand side of Eq. S5 is for constraining $\varepsilon$. The second term on the right-hand side of Eq. S5 is a regularization term. $C_\varepsilon^{-1}$ and $C_M^{-1}$ are used as weighting matrix for balancing the two terms on the right-hand side of Eq. S5. Supposing that no error will be incurred in calculating the partial derivatives in Eq. S4, the true PDE will have no prediction error, that is, $\varepsilon^{obs} = 0$.

Using the Gauss-Newton method for minimizing the objective function shown in Eq. S5, the iterative update formula takes the form:

$$m_{l+1} = m_l - \left(C_M^{-1} + G_l^T C_\varepsilon^{-1} G_l\right)^{-1}\left[C_M^{-1}\left(m_l - m^{pr}\right) + G_l^T C_\varepsilon^{-1}\left(\varepsilon(m_l) - \varepsilon^{obs}\right)\right], \qquad (S6)$$

where $l$ denotes the iteration index; and $G_l$ denotes the gradient taking value at $m_l$. In Eq. S6, $C_M^{-1} + G_l^T C_\varepsilon^{-1} G_l$ is the Hessian matrix without the term containing the second-order derivative of $O(m)$. In order to mitigate the influence of the large data mismatch in early iterations, the Hessian can be modified by using a multiplier $\lambda$ to $C_M^{-1}$ in the Hessian term as:

$$m_{l+1} = m_l - \left[(1+\lambda_l)C_M^{-1} + G_l^T C_\varepsilon^{-1} G_l\right]^{-1}\left[C_M^{-1}\left(m_l - m^{pr}\right) + G_l^T C_\varepsilon^{-1}\left(\varepsilon(m_l) - \varepsilon^{obs}\right)\right]. \qquad (S7)$$

Using the following equalities:

$$\left(C_M^{-1} + G_l^T C_\varepsilon^{-1} G_l\right)^{-1} = C_M - C_M G_l^T \left(C_\varepsilon + G_l C_M G_l^T\right)^{-1} G_l C_M, \qquad (S8)$$

$$\left(C_M^{-1} + G_l^T C_\varepsilon^{-1} G_l\right)^{-1} G_l^T C_\varepsilon^{-1} = C_M G_l^T \left(C_\varepsilon + G_l C_M G_l^T\right)^{-1}. \qquad (S9)$$

Eq. S7 can be rewritten as:

$$\begin{aligned} m_{l+1} = m_l \\ - \frac{1}{1+\lambda_l}\left[C_M - C_M G_l^T \left((1+\lambda_l)C_\varepsilon + G_l C_M G_l^T\right)^{-1} G_l C_M\right] C_M^{-1}\left(m_l - m^{pr}\right) \\ - C_M G_l^T \left((1+\lambda_l)C_\varepsilon + G_l C_M G_l^T\right)^{-1}\left(\varepsilon(m_l) - \varepsilon^{obs}\right). \end{aligned} \qquad (S10)$$

Eq. S10 is the data-assimilation method utilized in this work. For implementing the data-assimilation method, the gradient $G$ needs to be calculated. In this work, finite difference is utilized for calculating $G$ as:

$$G_{l,i} = \frac{\varepsilon(m_l + \Delta m_{l,i}) - \varepsilon(m_l - \Delta m_{l,i})}{2\Delta m_{l,i}}, \quad (S11)$$

where $G_{l,i}$ and $m_{l,i}$ denote the $i^{th}$ entry of $G_l$ and $m_l$, respectively; and $\Delta m_{l,i}$ denotes the perturbation of $m_{l,i}$. Here, note that, for calculating $G_{l,i}$, we only perturb the $i^{th}$ entry of $m_l$, keeping other entries unchanged. In this work, $\Delta m_{l,i}$ is set to be 1% of $m_{l,i}$.

After each update of $m$, $\varepsilon(m)$ will be calculated. If $\varepsilon(m_{l+1}) < \varepsilon(m_l)$, the update will be accepted and $\lambda$ will be reduced by a factor $\gamma$. Otherwise, the update will be rejected, and the $\lambda$ will be increased by a factor $\gamma$ to repeat the current iteration. In this work, $\gamma$ is set to be 10.

In summary, starting with one initial sample of $m$, $\alpha$ will be learned and $\varepsilon$ will be calculated. Then, the data-assimilation method can iteratively update $m$ to minimize an objective function. After each update of $m$, $\alpha$ will be relearned and $\varepsilon$ will be recalculated. This process will continue until a convergence criterion is met. The convergence criterion utilized in this work is given as:

(1) $\varepsilon(m_{l+1}) - \varepsilon(m_l) < \tau \times \varepsilon(m_l)$;

(2) Iteration exceeds the pre-given number, $I_{MAX}$.

In this work, we set $\tau = 0.001$ and $I_{MAX} = 25$.

### 1.1 Finite difference scheme for calculating derivatives

Derivative calculation is needed in this work both for building $\partial U/\partial t$ and $\Phi(U,m)$ (shown in Eq. S2) using the training data set and calculating $\varepsilon(m)$ (shown in Eq. S4) using the testing data set. In this work, the finite difference scheme for calculating derivatives takes the form:

$$\frac{\partial u}{\partial t}(x, t_k) = (0.5u(x, t_{k+1}) - 0.5u(x, t_{k-1}))/\Delta t,$$

$$\frac{\partial u}{\partial x}(x_i, t) = (0.5u(x_{i+1}, t) - 0.5u(x_{i-1}, t))/\Delta x,$$

$$\frac{\partial^2 u}{\partial x^2}(x_i, t) = (u(x_{i+1}, t) - 2u(x_i, t) + u(x_{i-1}, t))/(\Delta x)^2, \quad \text{(S12)}$$

$$\frac{\partial^3 u}{\partial x^3}(x_i, t) = (0.5u(x_{i+2}, t) - u(x_{i+1}, t) + u(x_{i-1}, t) - 0.5u(x_{i-2}, t))/(\Delta x)^3.$$

where $\Delta t$ and $\Delta x$ are the step sizes for time and space, respectively. Here, note that, to calculate the derivatives, we need the spatially or temporally nearby data. The data near the boundaries without sufficient nearby data for calculating the derivatives are not utilized for learning PDE.

**1.2 Pretreatment of noisy data**

Since finite difference is used for calculating derivatives in this work, small data noise may incur huge error in the calculated derivatives. Thus, pretreatment of the noisy data is requisite. In this work, the polynomial technique is utilized to smooth the noisy data. The procedure is given below:

1. For each monitoring location $x_0$, smooth the data along $t$ according to the following procedures:

    a) For each $t_k$, $k = 1, ..., n_t$, design an interval $[t_k - n^{CH}\Delta t, t_k + n^{CH}\Delta t]$. Generate $1 + N^{CH}$ Chebyshev interpolation points $t_i^{CH}$, $i = 1, ..., 1 + N^{CH}$.

    b) For each $t_i^{CH}$, $i = 1, ..., 1 + N^{CH}$, design an interval $[t_i^{CH} - n^{LS}\Delta t, t_i^{CH} + n^{LS}\Delta t]$. Calculate the smoothed value at the Chebyshev interpolation point, $u^{LS}(x_0, t_i^{CH})$ by performing a least squares regression with polynomial up to order $N^{LS}$ using the data inside of the designed interval.

    c) Calculate the smoothed value $u^{CH}(x_0, t_k)$ by performing Chebyshev interpolation using the values $u^{LS}(x_0, t_i^{CH})$, $i = 1, ..., 1 + N^{CH}$.

2. For each monitoring step $t_0$, smooth the data along $x$ using the procedures described in step 1.

3. Calculate the derivatives using finite difference.
4. Repeat the above three steps, if the calculated derivatives exhibit unreasonable fluctuation.

For the investigated problem, we set $N^{CH}=5$ and $N^{LS}=3$. Considering that the data density in the temporal domain and spatial domain is different, we set $n^{CH}=n^{LS}=120$ for smoothing the data along *t* and set $n^{CH}=n^{LS}=6$ for smoothing the data along *x*. Here, due to the fact that there are not sufficient data for smoothing near the boundaries in time and space, the derivatives near the boundary are not employed for learning PDE.

**1.3 Constraining the update of model parameter**

If there exist lower and upper bounds for the model parameter, a transformation technique can be utilized for constraining the update of model parameter. Let $m^{lower}$ and $m^{upper}$ denote the lower and upper bounds for *m*, respectively. The transformed parameter, *s*, is defined by the following log-transformation:

$$s = \ln\left(\frac{m - m^{lower}}{m^{upper} - m}\right). \tag{S13}$$

Then, the update is applied on *s* with gradient $G^s$ calculated as:

$$G_i^s = \frac{\partial \varepsilon}{\partial s_i} = \frac{\partial \varepsilon}{\partial m_i}\frac{\partial m_i}{\partial s_i} = G_i^m \frac{(m_i^{upper} - m_i)(m_i - m_i^{lower})}{m_i^{upper} - m_i^{lower}}, \tag{S14}$$

where $G_i^m$ is calculated using Eq. S11. After updating *s*, the original model parameter *m* is calculated as:

$$m = \frac{1}{2}(m^{upper} + m^{lower}) + \frac{1}{2}(m^{upper} - m^{lower})\left(\frac{\exp(s)-1}{\exp(s)+1}\right). \tag{S15}$$

In this work, the lower and upper bounds for *a* is 0.25 and 0.75, respectively, and the lower and upper bounds for $K_l$ is 30 and 150, respectively. When implementing the proposed method, the transformation is not utilized for the first time. If the updated model parameter violates the constraint, the proposed method will be implemented again with model parameter transformation.

## Section S2. Cases

In this work, we evaluate the proposed method by considering the problem of contaminant solute transport in subsurface formation. A benchmark problem of solute transport from MT3DMS software (*23*) is adopted here. It is one-dimensional advective-dispersive transport with nonlinear equilibrium-controlled sorption. The flow field is steady state. The initial and boundary conditions of the transport model are:

$$C(x,0) = 0$$

$$\left. -\theta D \frac{\partial C}{\partial x} + qC \right|_{x=0} = \begin{cases} f_0, & 0 < t < t_0 \\ 0, & t > t_0 \end{cases} \quad \text{(S16)}$$

$$\frac{\partial C}{\partial x}(\infty, t) = 0, \quad t > 0$$

The values for model parameters are set as: grid spacing $\Delta x = 0.16\ cm$; dispersivity $\alpha_L = 1\ cm$; porosity $\theta = 0.37$; bulk density $\rho_b = 1.587\ g/cm^3$; duration of source pulse $t_0 = 160\ s$; advective mass flux $f_0 = qC_0$ and concentration of source fluid $C_0 = 5 \times 10^{-2}\ mg/l$. Three solute transport scenarios are designed here: scenario 1 contains the two processes of ADV and DIS; scenario 2 contains the three processes of ADV, DIS, and F-SORP; and scenario 3 contains the three processes of ADV, DIS, and L-SORP. MT3DMS software is utilized for running the simulations to obtain the measurement data for the three scenarios. The solute concentration is measured at 101 evenly-distributed spatial locations. The concentration that is larger than $5 \times 10^{-5}\ mg/l$ is used as the data for identifying the physical processes.

**2.1 Cases with different parameter setups**

In order to test the robustness of the proposed method, here we design additional cases with different parameter setups.

In the first group of cases, scenario 2 and scenario 3 are reinvestigated with different values for $K_f$ and $K_l$, respectively. For scenario 2, $K_f$ is set to be 0.1 $(\mu g/g)(l/mg)^a$, and for scenario 3, $K_l$ is set to be 60 *l/mg*. The values of other parameters are unchanged as: $v_x = 0.01\ cm/s$; *a*=0.7 for scenario 2; and $\bar{S} = 0.003\ \mu g/g$ for scenario 3. The solute concentration is measured from 300 s to 1100 s, and the measurement interval is 0.5 s. Fig. S1 shows the learned coefficient of normalized terms and the estimation of model parameters. Table S1 presents the learned equations. Overall, the results are satisfactory.

In the second group of cases, scenario 2 and scenario 3 are reinvestigated with different values for $v_x$. We set $v_x = 0.05\, cm/s$ for the two scenarios. The values of other parameters are unchanged as: $K_f = 0.05\, (\mu g/g)(l/mg)^a$ for scenario 2; $a=0.7$ for scenario 2; $K_l = 100\, l/mg$ for scenario 3; and $\bar{S} = 0.003\, \mu g/g$ for scenario 3. Considering that the solute plume will rapidly move through the investigated domain due to the larger average linear groundwater velocity, the solute concentration should be measured at a shorter period with larger frequency. Here, the solute concentration is measured from 180 s to 300 s, and the measurement interval is 0.1 s. Fig. S2 shows the learned coefficient of normalized terms and the estimation of model parameters. Table S2 presents the learned equations. For scenario 2, one can obviously see the spurious update of Langmuir constant for some implementations. However, since the empirical model can be correctly selected, this spurious update of model parameter will not influence the final learned equation. The learned equations shown in Table S2 are satisfactory.

## 2.2 Results with data noise

In this work, for testing the performance of the proposed method, three noise levels, 1%, 5%, and 10%, are added to the data of the three solute transport scenarios. Fig. S3 shows the estimation of Freundlich exponent for the three scenarios with three data-noise levels. Fig. S4 presents the estimation of Langmuir constant exponent for the three scenarios with three data-noise levels. For scenario 1, as the noise level increases, spurious update occurs for a small number of implementations. For scenario 2 and scenario 3, as the noise level increases, the accuracy of the updated model parameter of the empirical model that is identified to be occurred (or dominated) decreases, and more implementations show spurious update of the model parameter of the empirical model that is identified to be not occurred (or dominated).

## 2.3 Evaluating the results based on prediction error

Since the gradient-based data-assimilation method is utilized in the proposed framework, multiple implementations with different initial samples of model parameter are recommended for avoiding getting stuck at the local minimums of the objective function. Considering that a prediction error of the learned model for the testing data will be calculated in the proposed method, we can evaluate the results from different implementations based on the calculated predictions errors. Fig. S5 shows the prediction error and the learned coefficients of PDE terms for the three scenarios with clean data. It can be seen that, with clean data, the variability of the prediction error and the learned coefficients are small. Table S3 presents the mean and the standard deviation (std) of the coefficients of the PDE terms, and one can see that the coefficient of variation (which equals the standard deviation divided by the mean) of the coefficients of the PDE terms are small. As discussed in the main text, data noise may incur larger variability in the results from different implementations, which will increase the difficulty for physical process identification and model parameter inference. This can be clearly seen from the results of identification of scenario 2 with three data-noise levels, as shown in Fig. 2. For further evaluation of the results, Figs. S6a-S6c show the prediction error for scenario 2 with three data-noise levels, and one can see that the learned models from some implementations have obviously higher prediction errors than others, which may be screened out. Then, we screen out the results from some implementations that have prediction errors larger than a certain value indicated by the red dashed lines shown in Figs. S6a-S6c. The coefficient of the normalized terms of the resting implementations are shown in Figs. S6d-S6f, and one can see that the variability of the coefficients obviously decreases, which can benefit the physical process identification and model parameter inference. However, the variability of the coefficients is still not sufficiently small, especially for the results shown in Fig. S6d. In order to further improve the accuracy of the learned model, the proposed method may be implemented again with a modified candidate library, which only retains the identified terms from the first implementation. For scenario 2 with three data-noise levels, since it can be identified that F-SORP is the better empirical model for SORP, the proposed method may be implemented again by deleting the candidate term for L-SORP in the candidate library. Fig. S7 shows the results of identification of scenario 2 with a modified candidate library, and it can be seen that the variability of the learned coefficients and the updated model parameters is small, which can benefit the determination of the learned equations. For the three scenarios with different data-noise levels, the mean and the standard deviation of the coefficients of the PDE terms are given in Table S3. The results are obtained by screening out the abnormal results and implementing the proposed method again with a modified candidate library if necessary. It can be seen that the coefficients of ADV, DIS, and L-SORP usually have a small coefficient of variation. However, the coefficient of F-SORP may have a larger coefficient of variation for large data-noise levels than that of other terms. Overall, the variability of the learned equations from different implementations is small.

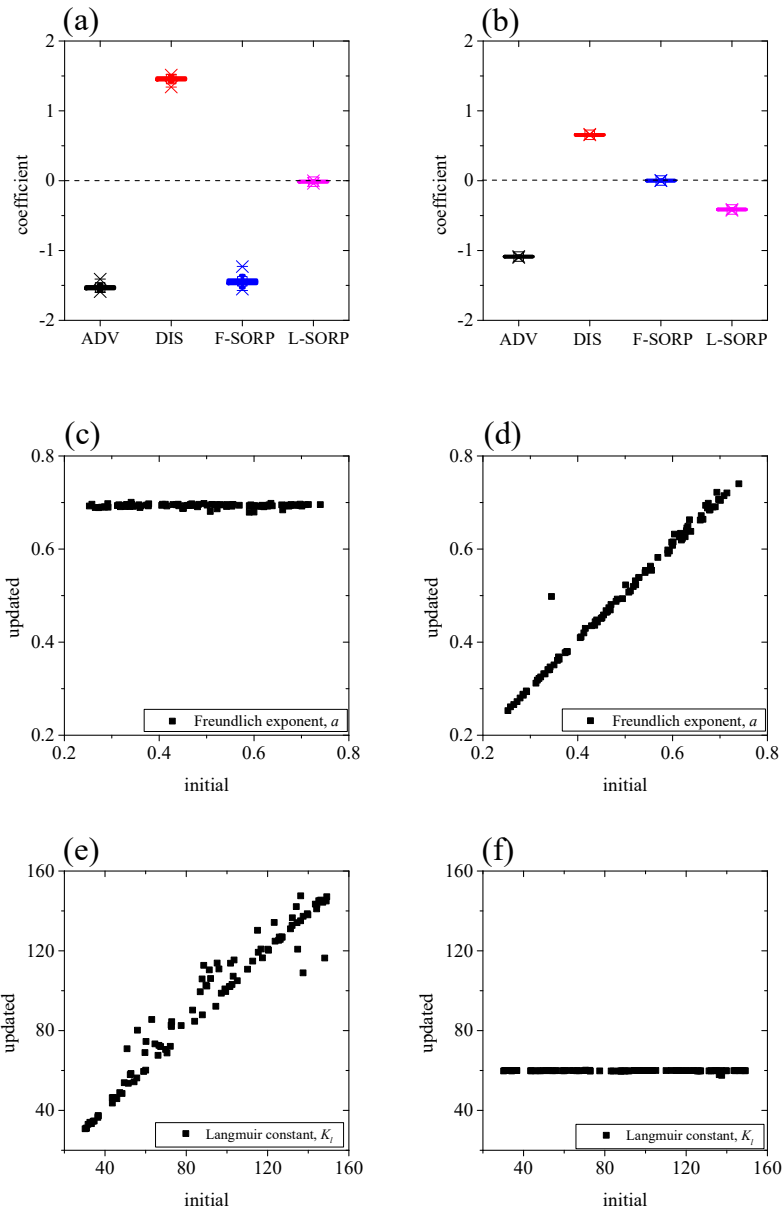

**Fig. S1. The results of identification of scenario 2 and scenario 3 with different values for $K_f$ and $K_l$, respectively.** The learned coefficient of normalized terms of scenario 2 (**a**) and scenario 3 (**b**). The estimation of Freundlich exponent of scenario 2 (**c**) and scenario 3 (**d**). The estimation of Langmuir constant of scenario 2 (**e**) and scenario 3 (**f**).

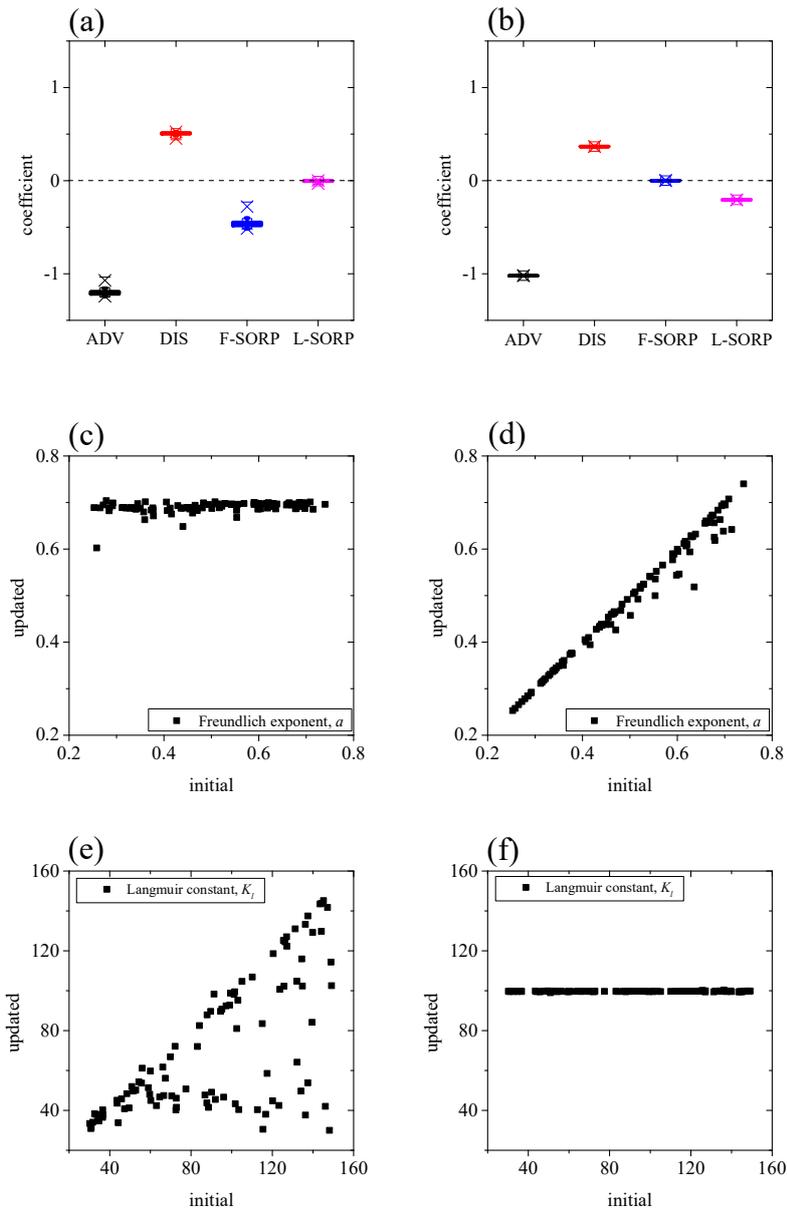

**Fig. S2. The results of identification of scenario 2 and scenario 3 with different value for $v_x$.** The learned coefficient of normalized terms of scenario 2 (**a**) and scenario 3 (**b**). The estimation of Freundlich exponent of scenario 2 (**c**) and scenario 3 (**d**). The estimation of Langmuir constant of scenario 2 (**e**) and scenario 3 (**f**).

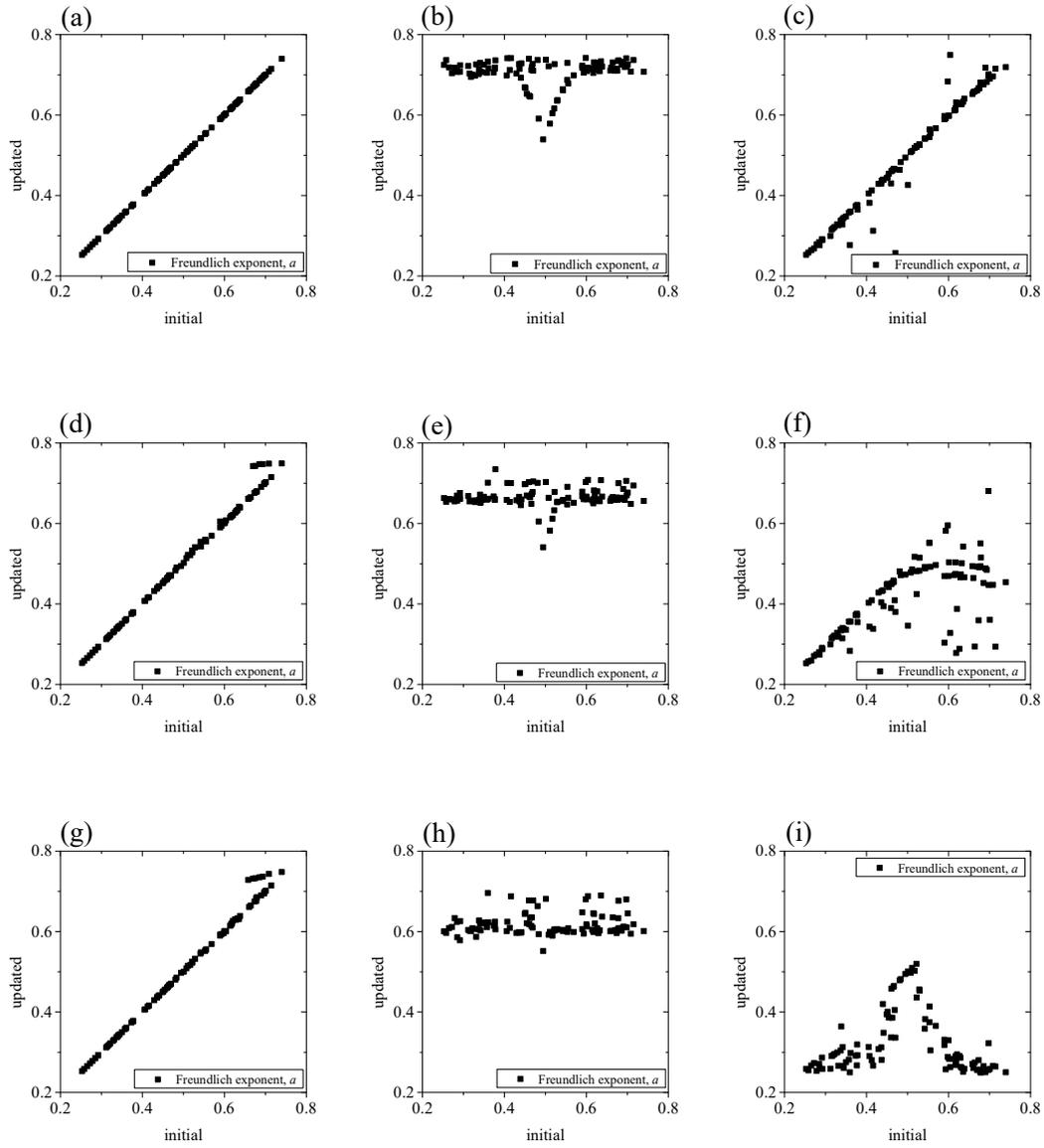

**Fig. S3. The estimation of Freundlich exponent for the three scenarios with three data-noise levels.** With 1% noise, the results of scenario 1 (**a**), scenario 2 (**b**), and scenario 3 (**c**). With 5% noise, the results of scenario 1 (**d**), scenario 2 (**e**), and scenario 3 (**f**). With 10% noise, the results of scenario 1 (**g**), scenario 2 (**h**), and scenario 3 (**i**).

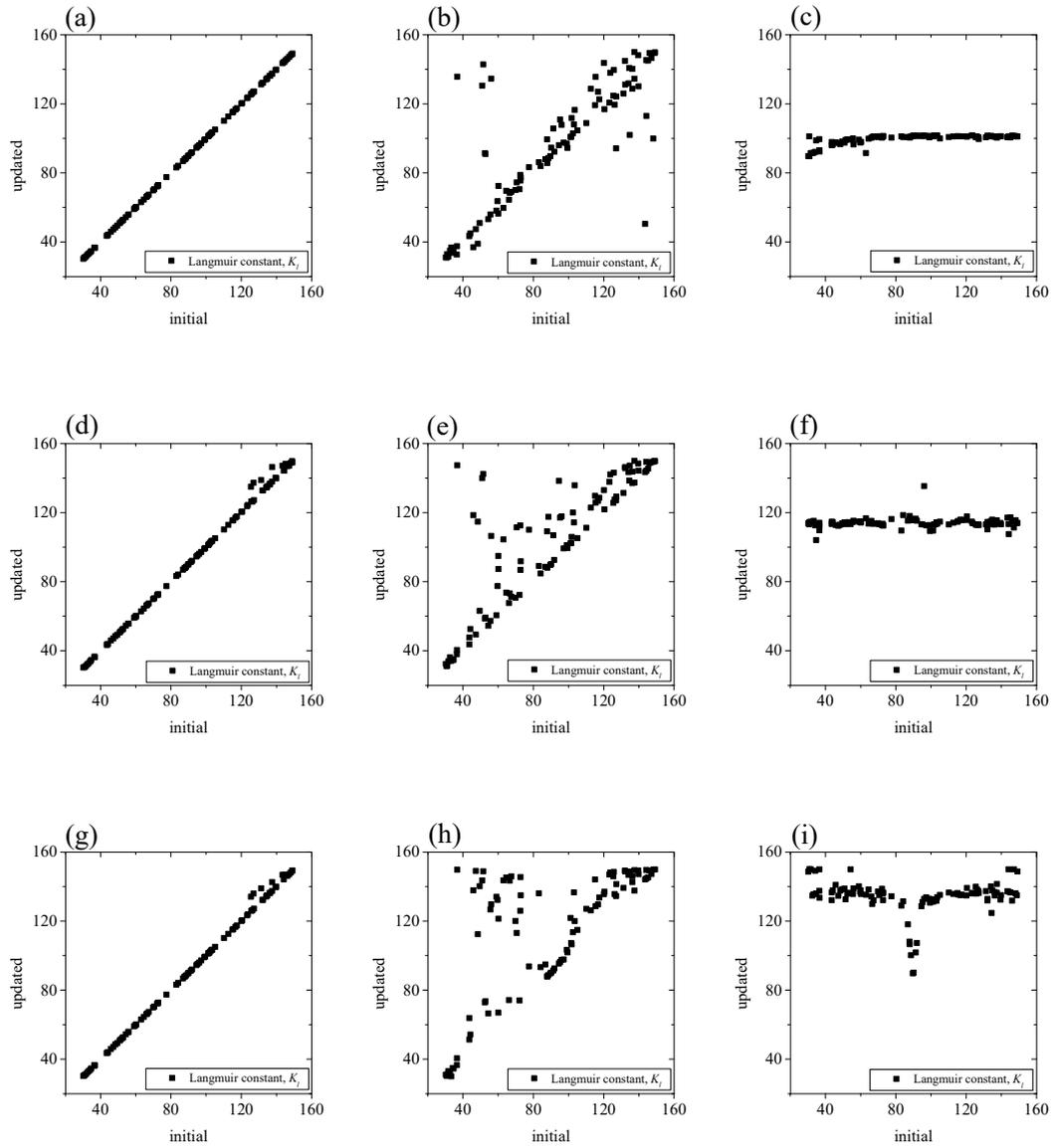

**Fig. S4. The estimation of Langmuir constant for the three scenarios with three data-noise levels.** With 1% noise, the results of scenario 1 (**a**), scenario 2 (**b**), and scenario 3 (**c**). With 5% noise, the results of scenario 1 (**d**), scenario 2 (**e**), and scenario 3 (**f**). With 10% noise, the results of scenario 1 (**g**), scenario 2 (**h**), and scenario 3 (**i**).

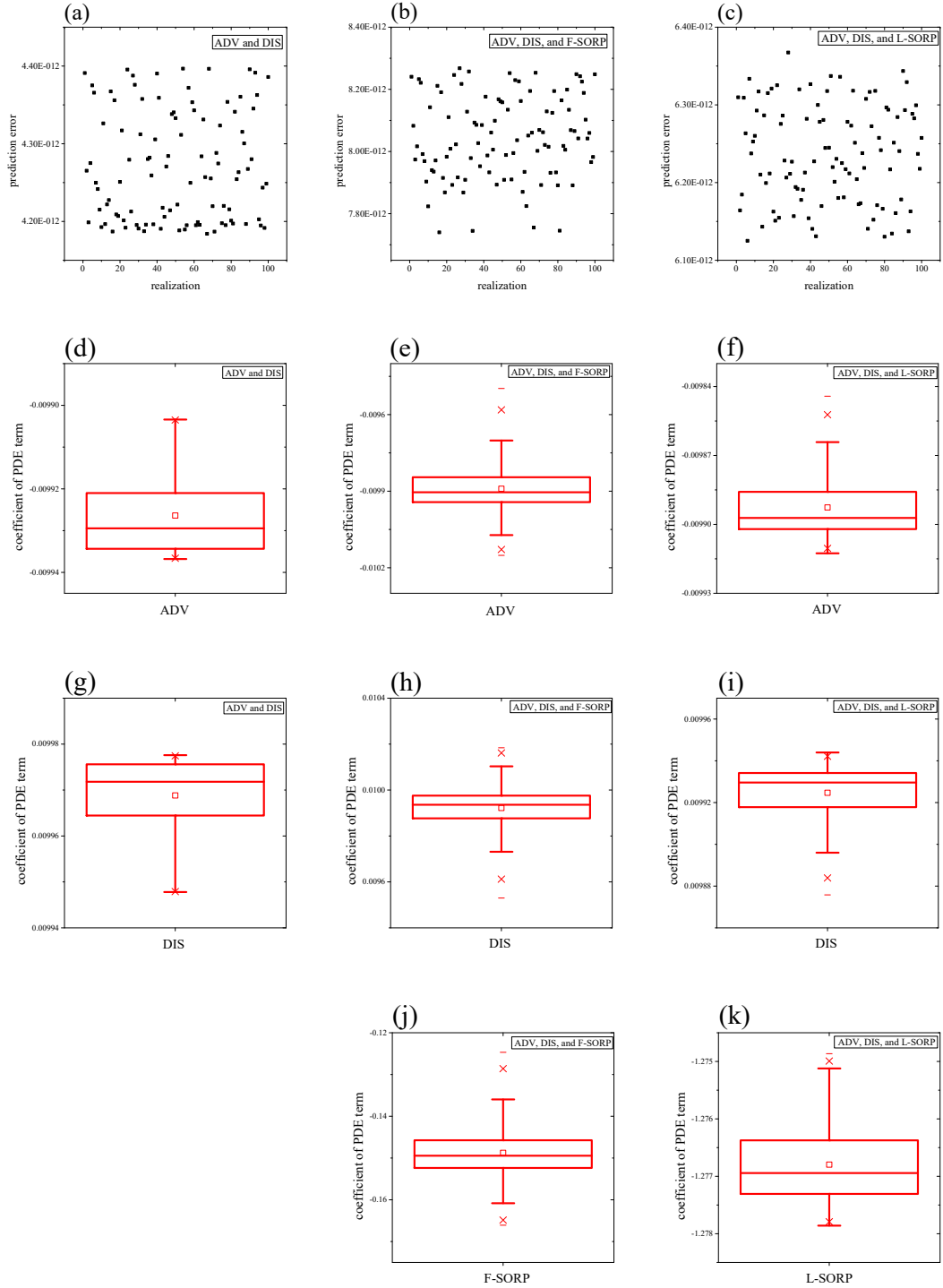

**Fig. S5. The prediction error and the learned coefficients of PDE terms for the three scenarios with clean data.** The prediction error of scenario 1 (**a**), scenario 2 (**b**), and scenario 3 (**c**). The learned coefficient of ADV of scenario 1 (**d**), scenario 2 (**e**), and scenario 3 (**f**). The learned coefficient of DIS of scenario 1 (**g**), scenario 2 (**h**), and scenario 3 (**i**). The learned coefficient of F-SORP of scenario 2 (**j**). The learned coefficient of L-SORP of scenario 3 (**k**).

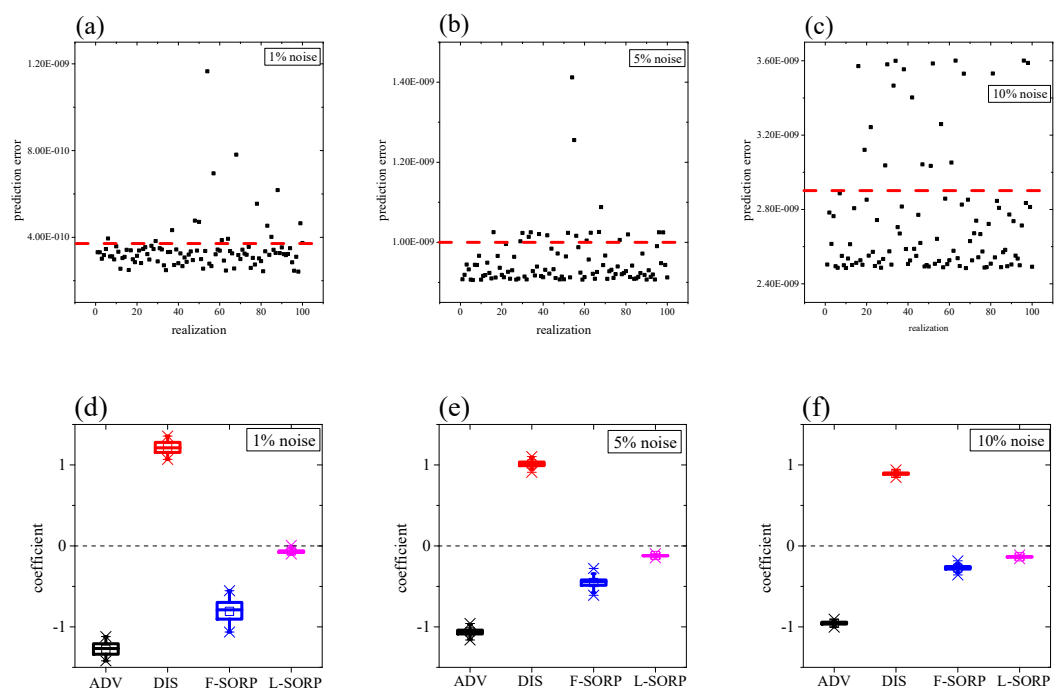

**Fig. S6. The prediction error and the learned coefficients of normalized terms for scenario 2 with three data-noise levels.** The prediction error with 1% noise (**a**), 5% noise (**b**), and 10% noise (**c**). The learned coefficient of normalized terms with 1% noise (**d**), 5% noise (**e**), and 10% noise (**f**).

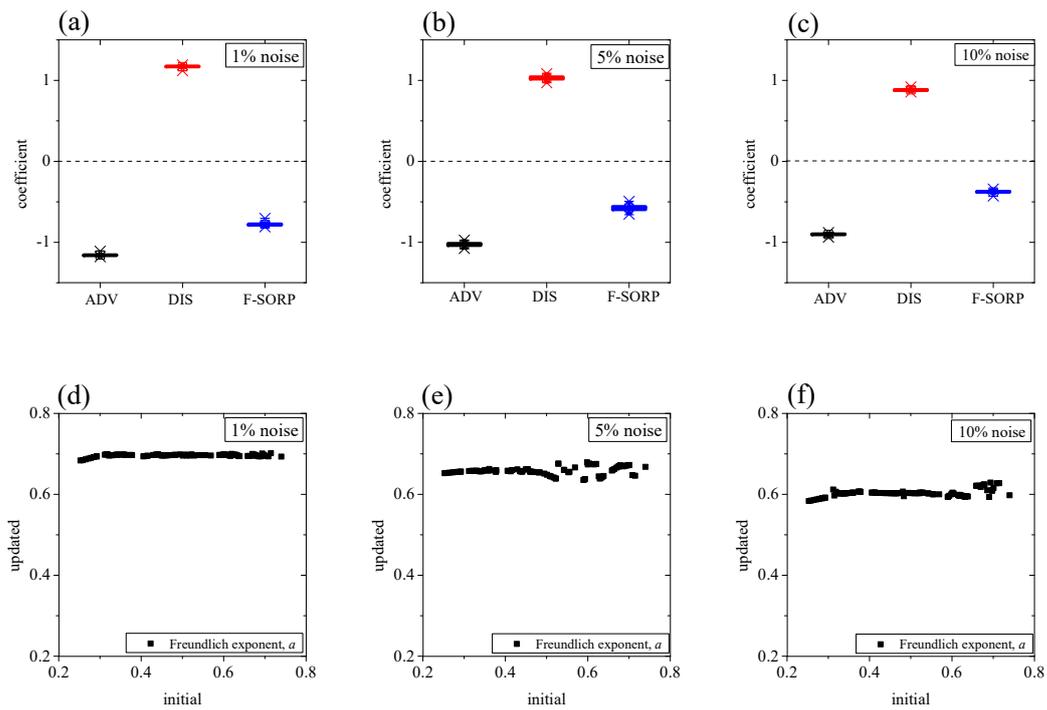

**Fig. S7. The results of identification of scenario 2 with a modified candidate library.** The learned coefficients of normalized terms with 1% noise (**a**), 5% noise (**b**), and 10% noise (**c**). The estimation of Freundlich exponent with 1% noise (**d**), 5% noise (**e**), and 10% noise (**f**).

**Table S1.** The learned equations of scenario 2 and scenario 3 with different values for $K_f$ and $K_l$, respectively.

| Physical processes | Learned equation |
|---|---|
| ADV, DIS, and F-SORP $$\frac{\partial C}{\partial t} = -0.01\frac{\partial C}{\partial x} + 0.01\frac{\partial^2 C}{\partial x^2} - 0.300 C^{0.7-1}\frac{\partial C}{\partial t}$$ | $$\frac{\partial C}{\partial t} = -0.00943\frac{\partial C}{\partial x} + 0.00945\frac{\partial^2 C}{\partial x^2} - 0.262 C^{0.693-1}\frac{\partial C}{\partial t} \quad \text{(clean data)}$$ |
| ADV, DIS, and L-SORP $$\frac{\partial C}{\partial t} = -0.01\frac{\partial C}{\partial x} + 0.01\frac{\partial^2 C}{\partial x^2} - 0.772\frac{1}{(1+60.0C)^2}\frac{\partial C}{\partial t}$$ | $$\frac{\partial C}{\partial t} = -0.00991\frac{\partial C}{\partial x} + 0.00995\frac{\partial^2 C}{\partial x^2} - 0.766\frac{1}{(1+59.899C)^2}\frac{\partial C}{\partial t} \quad \text{(clean data)}$$ |

**Table S2.** The learned equations of scenario 2 and scenario 3 with different value for $v_x$.

| Physical processes | Learned equation |
|---|---|
| ADV, DIS, and F-SORP $$\frac{\partial C}{\partial t} = -0.05\frac{\partial C}{\partial x} + 0.05\frac{\partial^2 C}{\partial x^2} - 0.150 C^{0.7-1}\frac{\partial C}{\partial t}$$ | $$\frac{\partial C}{\partial t} = -0.0488\frac{\partial C}{\partial x} + 0.0490\frac{\partial^2 C}{\partial x^2} - 0.134 C^{0.691-1}\frac{\partial C}{\partial t} \quad \text{(clean data)}$$ |
| ADV, DIS, and L-SORP $$\frac{\partial C}{\partial t} = -0.05\frac{\partial C}{\partial x} + 0.05\frac{\partial^2 C}{\partial x^2} - 1.287\frac{1}{(1+100.0C)^2}\frac{\partial C}{\partial t}$$ | $$\frac{\partial C}{\partial t} = -0.0499\frac{\partial C}{\partial x} + 0.0501\frac{\partial^2 C}{\partial x^2} - 1.284\frac{1}{(1+99.691C)^2}\frac{\partial C}{\partial t} \quad \text{(clean data)}$$ |

**Table S3.** The mean and standard deviation of the coefficients of the PDE terms.

| Physical processes | Data quality | Coefficient of ADV | | Coefficient of DIS | | Coefficient of F-SORP | | Coefficient of L-SORP | |
|---|---|---|---|---|---|---|---|---|---|
| | | mean | std | mean | std | mean | std | mean | std |
| ADV and DIS | clean data | -9.93E-3 | 9.33E-6 | 9.97E-3 | 8.25E-6 | | | | |
| | 1% noise | -9.97E-3 | 1.39E-5 | 1.002E-2 | 1.34E-5 | | | | |
| | 5% noise | -9.97E-3 | 1.56E-5 | 1.006E-2 | 1.73E-5 | | | | |
| | 10% noise | -9.91E-3 | 4.81E-5 | 1.004E-2 | 5.75E-5 | | | | |
| ADV, DIS, and F-SORP | clean data | -9.89E-3 | 9.75E-5 | 9.92E-3 | 9.79E-5 | -1.49E-1 | 6.58E-3 | | |
| | 1% noise | -9.67E-3 | 4.03E-5 | 9.68E-3 | 4.24E-5 | -1.38E-1 | 2.27E-3 | | |
| | 5% noise | -8.51E-3 | 6.33E-5 | 8.45E-3 | 6.82E-5 | -7.91E-2 | 3.13E-3 | | |
| | 10% noise | -7.50E-3 | 4.05E-5 | 7.26E-3 | 4.49E-5 | -3.68E-2 | 1.62E-3 | | |
| ADV, DIS, and L-SORP | clean data | -9.89E-3 | 1.28E-5 | 9.92E-3 | 1.29E-5 | | | -1.277E0 | 6.11E-4 |
| | 1% noise | -9.76E-3 | 6.03E-5 | 9.83E-3 | 6.11E-5 | | | -1.203E0 | 4.53E-3 |
| | 5% noise | -9.21E-3 | 5.09E-5 | 9.29E-3 | 5.23E-5 | | | -1.124E0 | 6.33E-3 |
| | 10% noise | -8.57E-3 | 7.88E-5 | 8.63E-3 | 8.27E-5 | | | -1.015E0 | 8.61E-3 |